\documentclass{article}
\usepackage{spconf,amsmath,graphicx}

\title{CROSS-ARCHITECTURE UNIVERSAL FEATURE CODING \\VIA DISTRIBUTION ALIGNMENT}
\address{anonymous}
\name{Changsheng Gao\textsuperscript{1}, Shan Liu\textsuperscript{2}, Feng Wu\textsuperscript{3}, and Weisi Lin\textsuperscript{1}
\address{
\textsuperscript{1}Nanyang Technological University, \textsuperscript{2}Tencent Media Lab, \textsuperscript{3}University of Science and Technology of China\\
\{changsheng.gao, wslin\}@ntu.edu.sg, shanl@global.tencent.com, fengwu@ustc.edu.cn}
\thanks{We acknowledge the support of GPU cluster built by MCC Lab of Information Science and Technology Institution, USTC. \emph{Corresponding author: Weisi Lin}}}

\begin{document}
%
\maketitle
\begin{abstract}
Feature coding has become increasingly important in scenarios where semantic representations rather than raw pixels are transmitted and stored. 
However, most existing methods are architecture-specific, targeting either CNNs or Transformers. This design limits their applicability in real-world scenarios where features from both architectures coexist. To address this gap, we introduce a new research problem: cross-architecture universal feature coding (CAUFC), which seeks to build a unified codec that can effectively compress features from heterogeneous architectures. 
To tackle this challenge, we propose a two-step distribution alignment method. First, we design the format alignment method that unifies CNN and Transformer features into a consistent 2D token format. Second, we propose the feature value alignment method that harmonizes statistical distributions via truncation and normalization. As a first attempt to study CAUFC, we evaluate our method on the image classification task. Experimental results demonstrate that our method achieves superior rate-accuracy trade-offs compared to the architecture-specific baseline.
This work marks an initial step toward universal feature compression across heterogeneous model architectures.
\end{abstract}

\begin{figure*}
    \centering
    \includegraphics[width=0.8\linewidth]{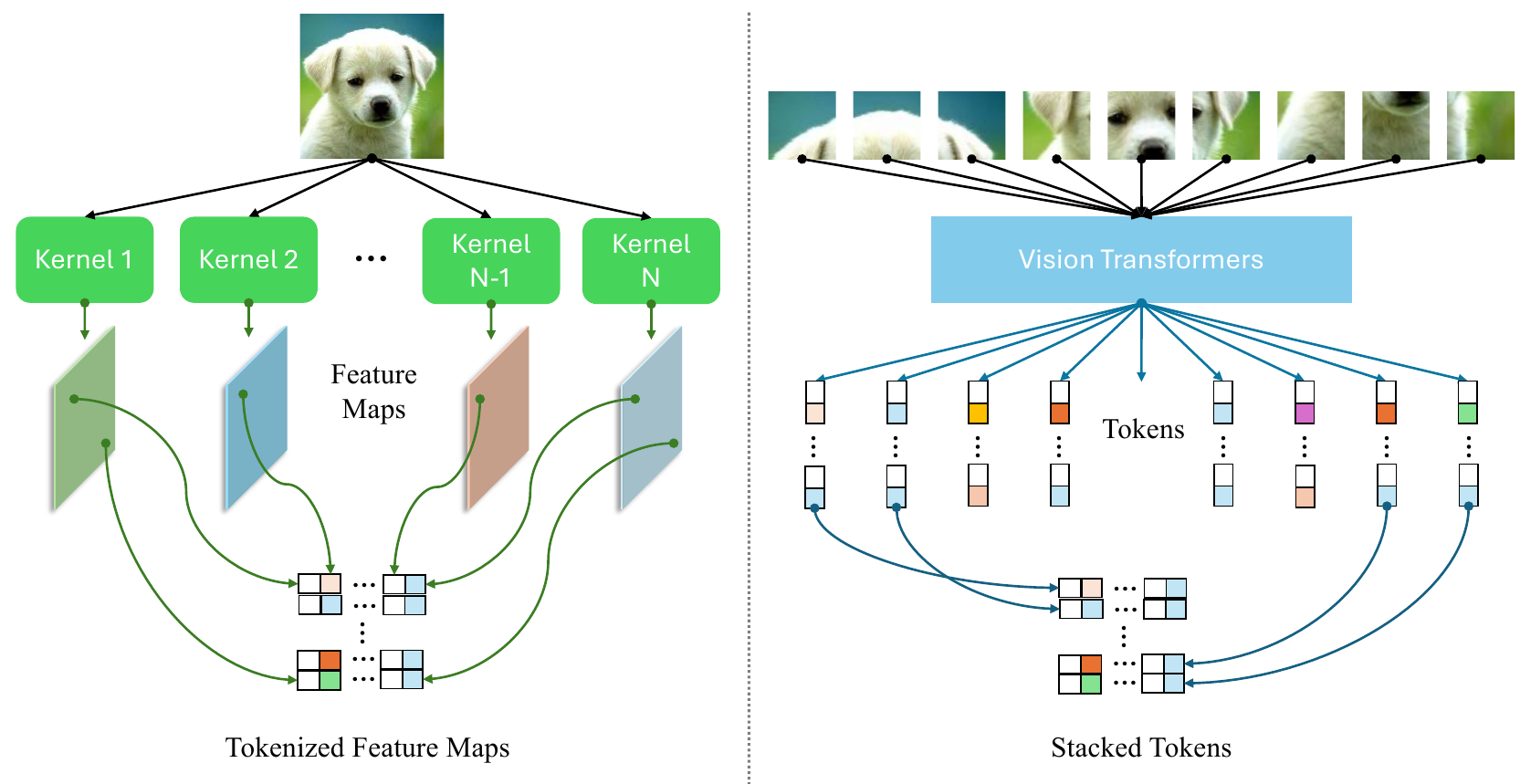}
    \caption{Illustration of feature extraction and format alignment for CNN features (left) and  Transformer features (right). The CNN model produces feature maps in the shape of $N\times H\times W$ while the Vision Transformer generates $M$ sequential $1\times L$ tokens. The CNN and Transformer features are aligned into a 2D format. (Refer to Sec. \ref{subsec_format_alignment} for more details.)}
    \label{fig_format_align}
\end{figure*}
\section{Introduction}
\label{sec:intro}
In recent years, coding for machines has received growing attention, which can be categorized into two branches: pixel coding and feature coding. Pixel coding compresses and reconstructs the original pixel data for downstream tasks \cite{choi2022scalable,tian2023nonsemantics,shen2024image,huang2025HMFVC}, while feature coding \cite{misra2022video,alvar2019multi,alvar2020bit,guo2023toward} focuses on compressing and reconstructing intermediate features extracted from images, which serve as semantic representations for further analysis.

Early attempts at feature coding leveraged handcrafted compression techniques \cite{fcm_cfp,choi2018deep,duan2016cdvs}, but these methods suffered from suboptimal performance due to distribution mismatches between images and features. Subsequently, researchers explored learned feature coding approaches that leverage the powerful distribution modeling capability of neural networks \cite{zhu2024learned,gao2024imofc,zhang2021MSFC,kim2023end}. 
More recently, Transformer-based large models have achieved remarkable success across various tasks. These models generate high-dimensional, semantically rich features that differ substantially from those produced by CNNs. As a result, feature coding for large models has attracted increasing interest \cite{oquab2023dinov2,guo2025deepseek}. Recent efforts \cite{gao2024feature} have built benchmarks and proposed codecs tailored to such features, highlighting the importance and uniqueness of large model feature coding.

Despite the extensive literature on feature coding, most existing methods are architecture-specific, targeting either convolutional neural networks (CNNs) or Vision Transformers (ViTs). However, real-world applications often involve a mixture of both architectures. \textbf{To address this, we introduce a new research direction in feature coding: cross-architecture universal feature coding (CAUFC)}. The objective is to design a unified codec capable of compressing features from both CNNs and Transformers.

One of the key challenges in CAUFC is the alignment of heterogeneous feature formats and their distinct statistical distributions. CNN features typically have a 3D structure composed of multiple 2D feature maps, while Transformer features are arranged as 2D token sequences lacking an explicit channel dimension. Additionally, the value ranges of CNN and Transformer features differ significantly.

In this paper, we study CAUFC in the context of the image classification task. We propose a two feature alignment strategies to provide the codec with consistent input representations. First, we perform format alignment to convert features into a unified 2D structure. Second, we perform value alignment to map all features into a consistent feature value range. 
Experimental results demonstrate that our codec achieves superior rate-accuracy trade-offs compared to the existing baseline.
Our contributions are summarized as follows:
\begin{itemize}
    \item We analyze the feature extraction mechanisms of CNNs and Transformers, and propose a format alignment method that unifies their outputs into a consistent 2D token representation.
    \item We investigate the statistical disparities between CNN and Transformer features, and introduce a feature value alignment method to harmonize their distributions.
    \item We validate the proposed CAUFC method on the image classification task using ResNet-50 and DINOv2 features, demonstrating superior rate-accuracy trade-offs compared to the existing baseline.
\end{itemize}

\begin{figure*}[htb]
    \centering
    \begin{minipage}[b]{0.24\linewidth}
        \centering
        \includegraphics[width=\linewidth]{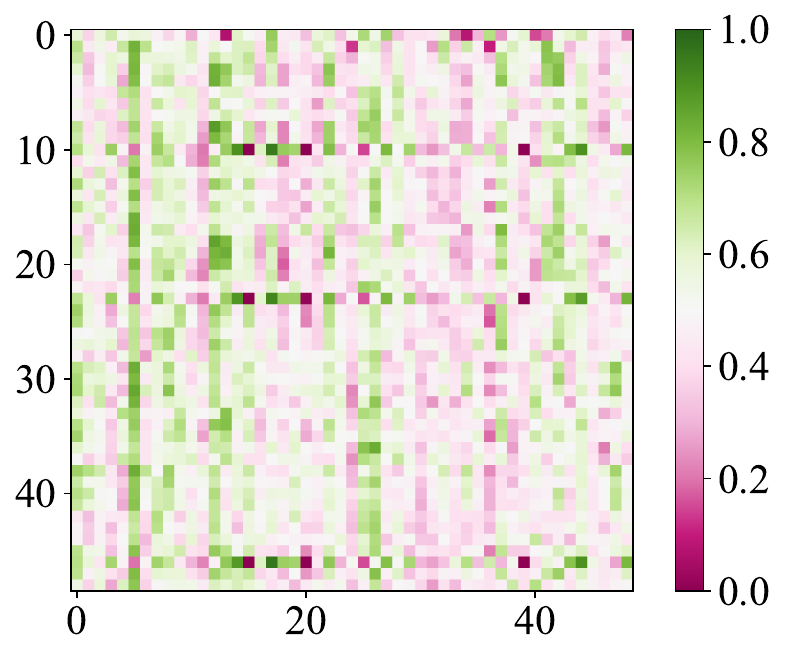}
        \centerline{(a)}\medskip
    \end{minipage}
    \begin{minipage}[b]{0.24\linewidth}
        \centering
        \includegraphics[width=\linewidth]{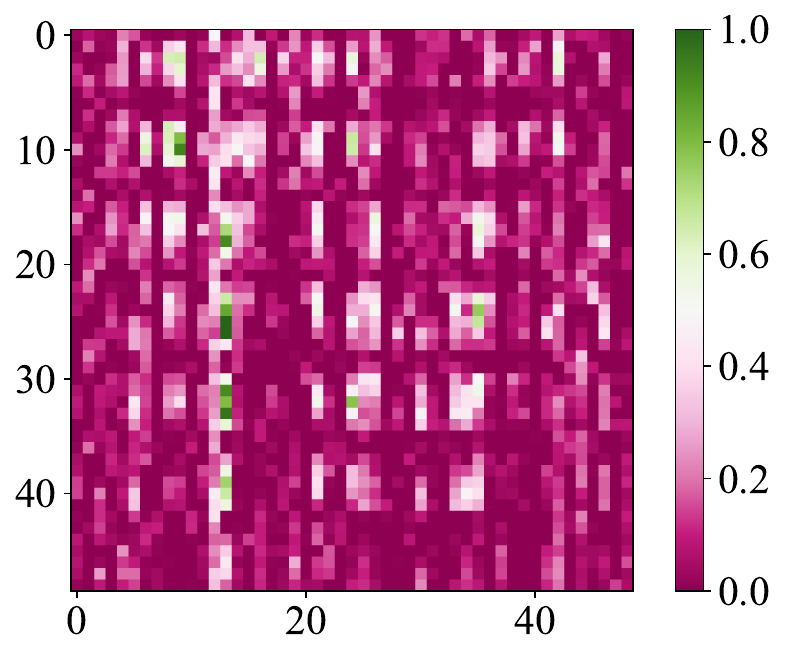}
        \centerline{(b)}\medskip
    \end{minipage}
    \begin{minipage}[b]{0.24\linewidth}
        \centering
        \includegraphics[width=\linewidth]{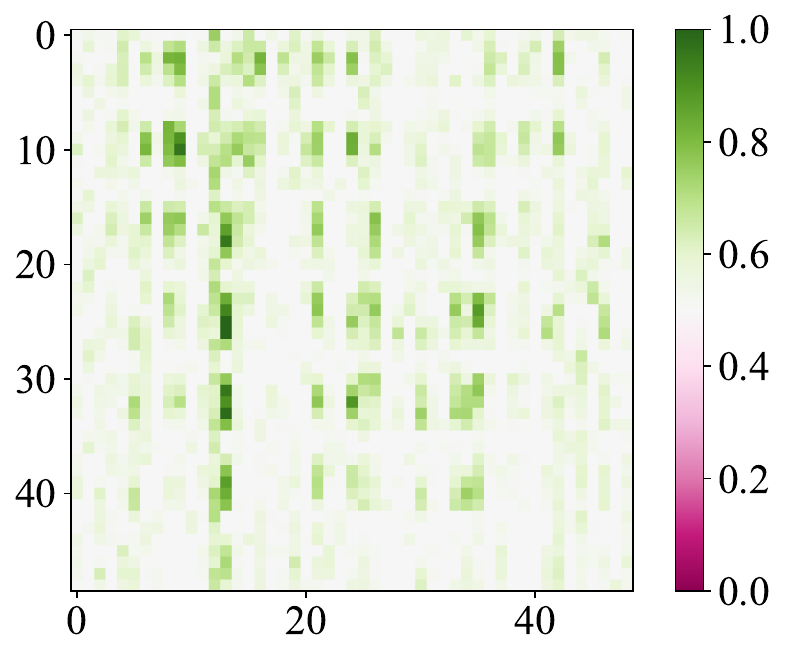}
        \centerline{(c)}\medskip
    \end{minipage}
    \caption{Feature value visualization of standard normalized DINOv2 feature (a), standard normalized ResNet feature (b), and shifted normalized ResNet feature (c).}
    \label{fig_value_visualization}
\end{figure*}

\section{The Proposed Method}
\label{sec_method}
\subsection{Feature Format Alignment}
\label{subsec_format_alignment}
We begin by introducing the feature extraction mechanisms in CNNs and Transformers, as illustrated in Fig.~\ref{fig_format_align}.
In CNNs, the entire input image $\mathbf{X}$ is processed through a series of convolutional layers. Each layer contains $N$ learnable kernels $\mathbf{K}_i$, each of which performs a localized weighted sum over a receptive field. This operation is applied across the spatial dimensions, producing a feature map of size $H \times W$ per kernel. After applying all kernels, the resulting feature tensor has shape $N \times H \times W$.

In contrast, ViTs split the image $\mathbf{X}$ into $M$ non-overlapping patches. Each patch $\mathbf{P}_i$ is projected into a $1 \times L$ token vector $\mathbf{T}_i$, which is then processed by a series of Transformer layers. Each layer updates these tokens via self-attention and outputs $M$ sequential token embeddings of shape $1 \times L$.

Since learned codecs require input features in an identical format, we convert both CNN and ViT features into a unified 2D format. ViT features are naturally arranged as $M \times L$ token matrices. For CNN features, we propose a tokenization strategy where the tensor is reshaped into $(H \times W) \times N$. Specifically, for each spatial location, we collect the values across all $N$ channels into a single $1 \times N$ token. Thus, the $H \times W$ spatial locations are transformed into $H \times W$ tokens.

We do not adopt the flattening method used in \cite{gao2024feature}, since it disrupts the semantic structure of CNN features. In ViTs, each dimension of the token corresponds to a specific learned attribute. Similarly, in CNNs, each kernel/channel is associated with a distinct semantic attribute. Therefore, sampling one attribute from each channel to construct a token ensures semantic alignment with ViT features. Furthermore, each token corresponds to a distinct spatial location, preserving positional consistency.

In our experiments, we focus on the image classification task. For CNNs, we use the output of the ResNet-50 C5 layer (denoted ResNet features). The original ResNet features are in the shape of $2048\times 7 \times 7$. For Transformers, we follow the setting in \cite{gao2024feature} and adopt features from the 40th decoder layer of DINOv2 (denoted DINOv2 features). DINOv2 features share a shape of $257\times 1536$. After format alignment, ResNet features and DINOv2 features are formatted into $49\times 2048$ and $257\times 1536$, respectively.

To demonstrate the reasonability of the proposed format alignment method, we visualize the aligned features in Fig. \ref{fig_value_visualization}. It can be seen that both the DINOv2 feature and the ResNet feature exhibit a strong vertical redundancy. This is because each column denotes the same attribute learned from the same ViT and CNN kernel.

\subsection{Feature Value Alignment}
\label{subsec_value_alignment}
\begin{figure*}[htb]
    \centering
    \begin{minipage}[b]{0.3\linewidth}
        \centering
        \includegraphics[width=\linewidth]{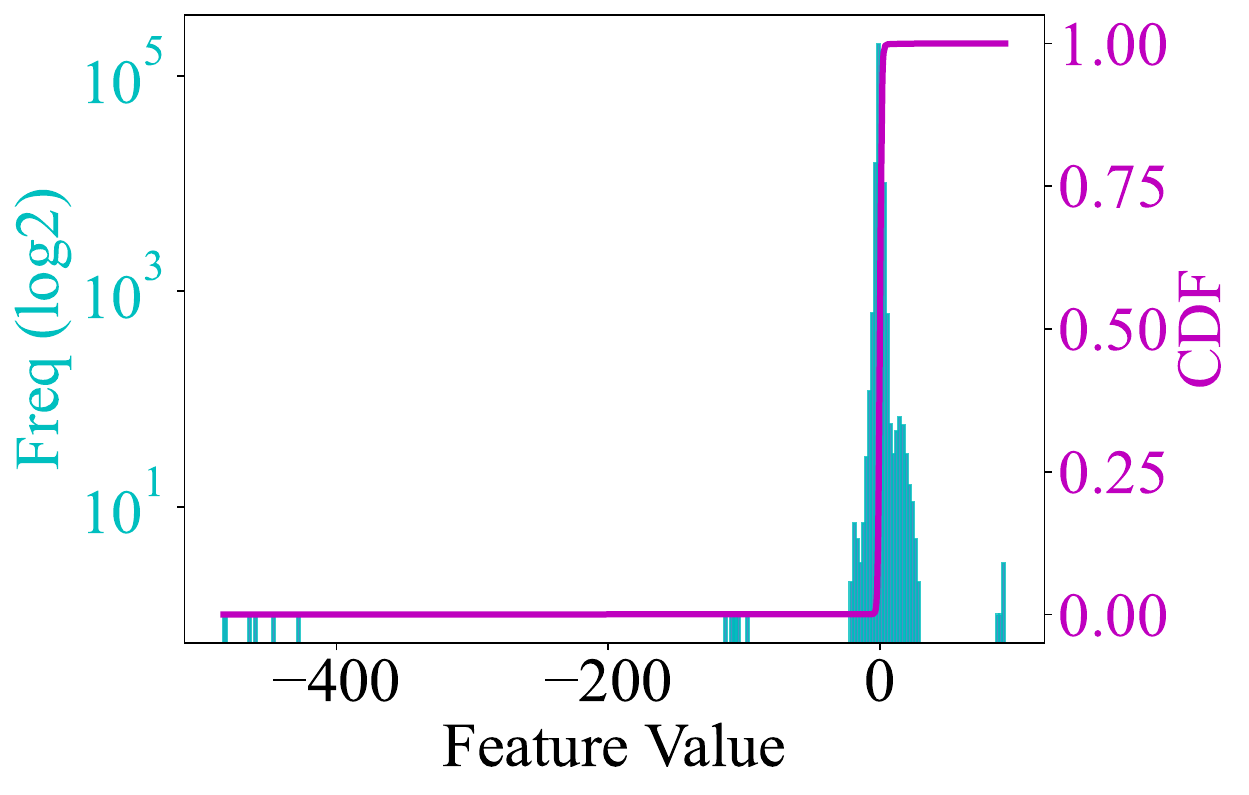}
    \end{minipage}
    \begin{minipage}[b]{0.3\linewidth}
        \centering
        \includegraphics[width=\linewidth]{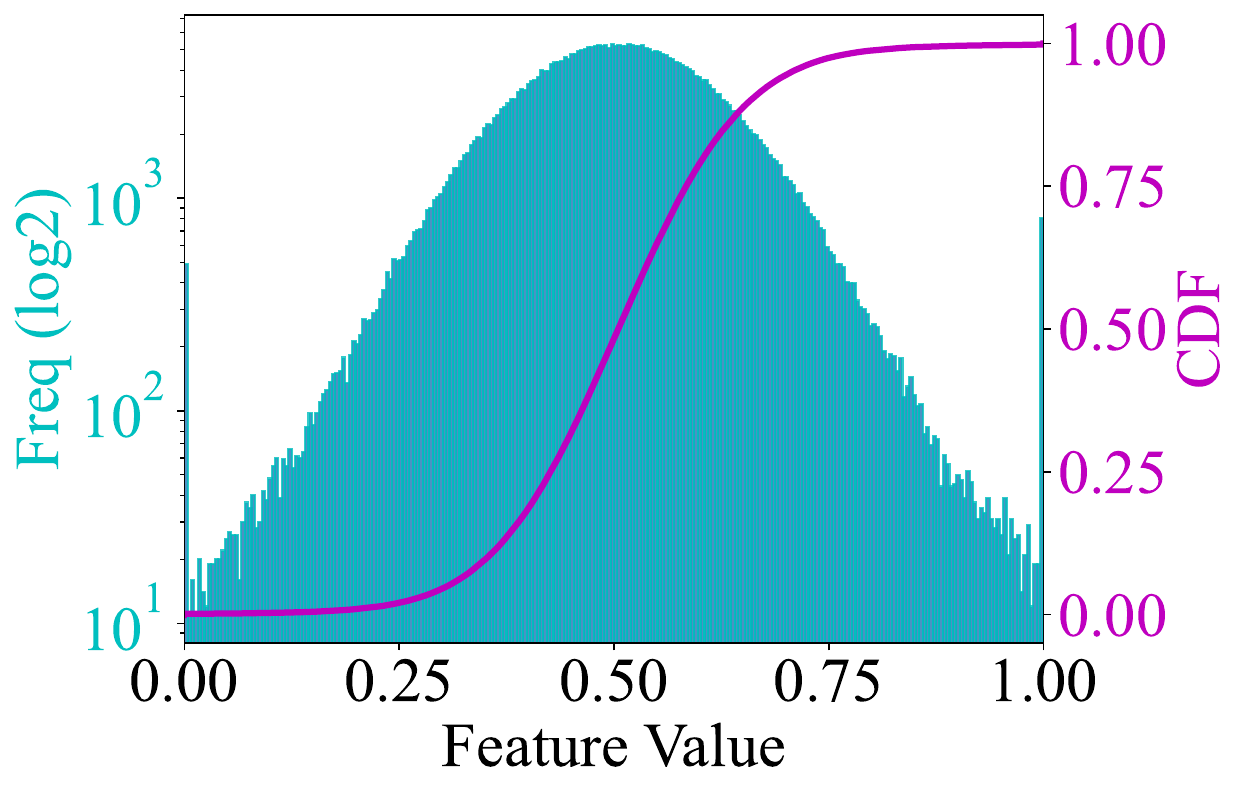}
    \end{minipage}
    \begin{minipage}[b]{0.3\linewidth}
        \centering
        \includegraphics[width=\linewidth]{figs/visualize/dinov2_n03661043_7216_uniquant_-5_5.pdf}
    \end{minipage}

    \begin{minipage}[b]{0.3\linewidth}
        \centering
        \includegraphics[width=\linewidth]{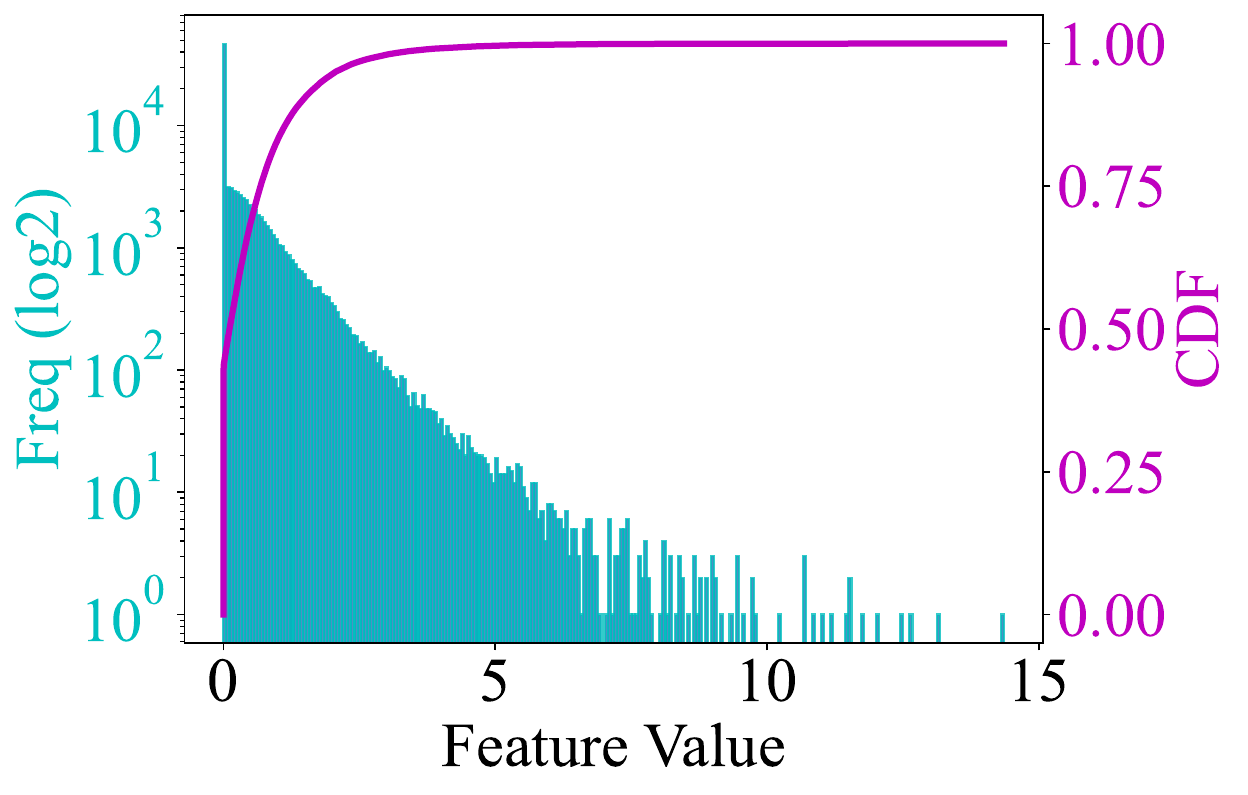}
    \end{minipage}
    \begin{minipage}[b]{0.3\linewidth}
        \centering
        \includegraphics[width=\linewidth]{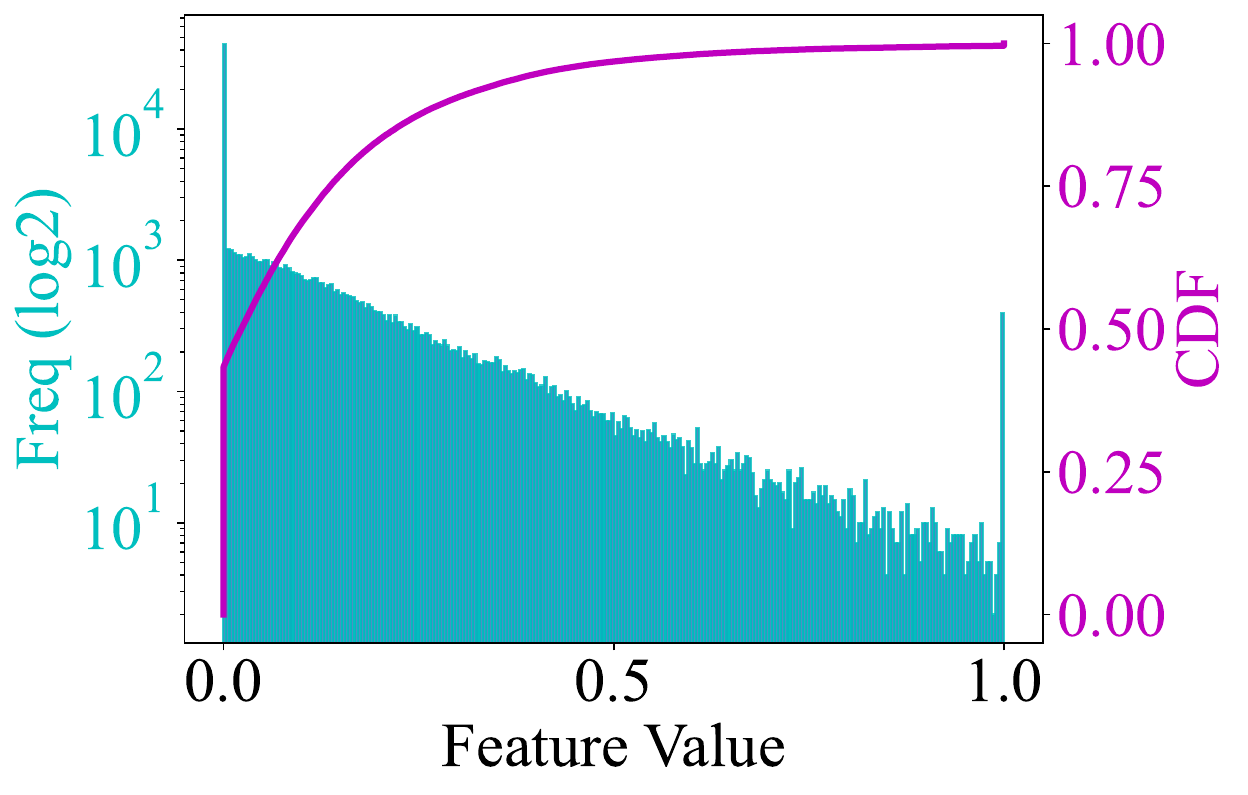}
    \end{minipage}
    \begin{minipage}[b]{0.3\linewidth}
        \centering
        \includegraphics[width=\linewidth]{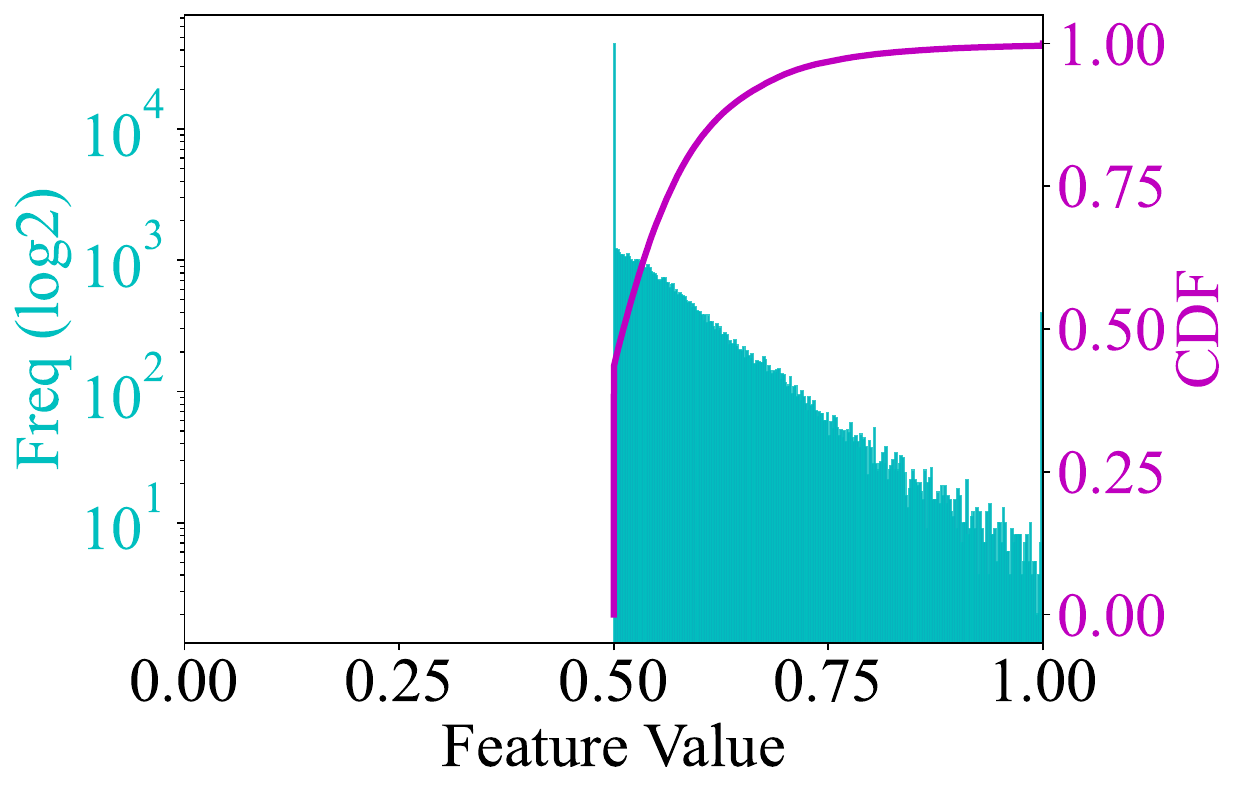}
    \end{minipage}

    \caption{Frequency and CDF visualization of the original features (left), standard normalized features (middle) and shifted normalized features (right). The top and bottom correspond to DINOv2 and ResNet features, respectively. The standard normalized DINOv2 feature is repeated (top right) to compare with the ResNet feature.}
    \label{fig_cdf}
\end{figure*}

We visualize the value distributions and cumulative density functions (CDFs) of ResNet and DINOv2 features in Fig.~\ref{fig_cdf}. DINOv2 features span a wide range but are concentrated within a narrow interval. ResNet features are strictly positive due to ReLU activations, with a narrower overall range. 

We propose a two-stage value alignment procedure: truncation and normalization. The truncation stage clips extreme values to ensure consistency and suppress outliers, transforming peaky distributions into balanced ones. A balanced input distribution facilitates the codec training because it makes the modeling of input distribution easier. We truncate DINOv2 and ResNet features to the ranges [-5, 5] and [0, 5], respectively. In our evaluation, the proposed truncation operation causes little performance drop in the image classification task. 

Next, we normalize the truncated features to the [0, 1] interval using
$x_{norm} = (x - lower) / (upper - lower)$,
where $lower$ and $upper$ are predefined lower and upper feature value bounds.
For DINOv2 features, we apply standard normalization by directly using their minimum and maximum values as the $lower$ and $upper$, respectively. For ResNet features, we propose a shifted normalization approach where the $lower$ is set to match the minimum value of DINOv2 features. We compare the features processed by these two normalization methods in Fig. \ref{fig_value_visualization} and Fig. \ref{fig_cdf}.
As shown in Fig. \ref{fig_value_visualization}, the shifted normalization makes ResNet features look more like DINOv2 features than standard normalization.
Fig. \ref{fig_cdf} reveals that while standard normalization maintains the same distribution range for both features, it results in different dominant regions. DINOv2 features predominantly fall within the [0.3, 0.7] range, whereas ResNet features are concentrated in [0, 0.3], with nearly half of the samples being zero. This discrepancy in dominant regions may potentially confuse the compression model during training. In contrast, the shifted normalization aligns ResNet features well with the right hand of DINOv2's distribution, despite not achieving exact numerical equivalence.

\begin{figure}
    \centering
    \includegraphics[width=0.7\linewidth]{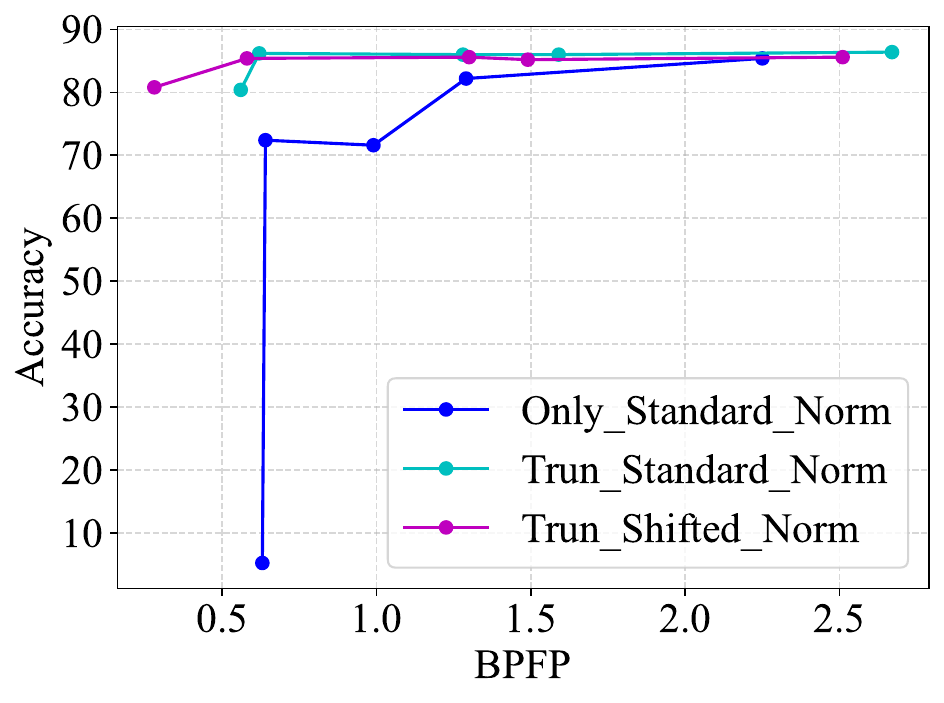}
    \caption{Rate-accuracy performance comparison between three feature alignment methods. (The baseline Hyperprior models are used here.)}
    \label{fig_rd_baseline}
\end{figure}

To verify the effectiveness of feature value alignment, we compare them in Fig. \ref{fig_rd_baseline}. We compare three methods: standard normalization alone, truncation + standard normalization, and truncation + shifted normalization. It is verified that truncation is necessary and shifted normalization exhibits higher accuracy at low bitrates. 

\subsection{Codec Training}
We adopt the Hyperprior model \cite{balle2018variational} as our feature codec. To adapt this architecture for 2D feature coding, we configure its input and output channels to 1.
During training, we construct the dataset by mixing DINOv2 and ResNet features. To balance the codec’s ability to handle both feature types, we adjust their sampling ratios in the training data. Specifically, at high bitrates, we assign a higher sampling ratio to DINOv2 features. This is motivated by prior work \cite{gao2024feature}, which shows that DINOv2 features exhibit lower spatial redundancy and are more challenging to model in the analysis transform. By increasing their representation in training, we enhance the codec’s ability to capture their distribution. At low bitrates, we reduce the sampling ratio for DINOv2 features, prioritizing ResNet features instead. This adjustment is made because (1) ResNet features contain simpler redundancies that are easier to compress at low bitrates, and (2) the limited information capacity at low bitrates makes it difficult for the codec to effectively model the complex distribution of DINOv2 features.

The feature codec is optimized using the following loss function
$L = BPFP + \lambda \times ||X - \hat{X}||^2$,
where BPFP (bits per feature point) measures the bitrate and $\lambda$ controls the trade-off between bitrate and feature distortion.
\begin{figure*}
    \centering
    \begin{minipage}[b]{0.35\linewidth}
        \centering
        \includegraphics[width=\linewidth]{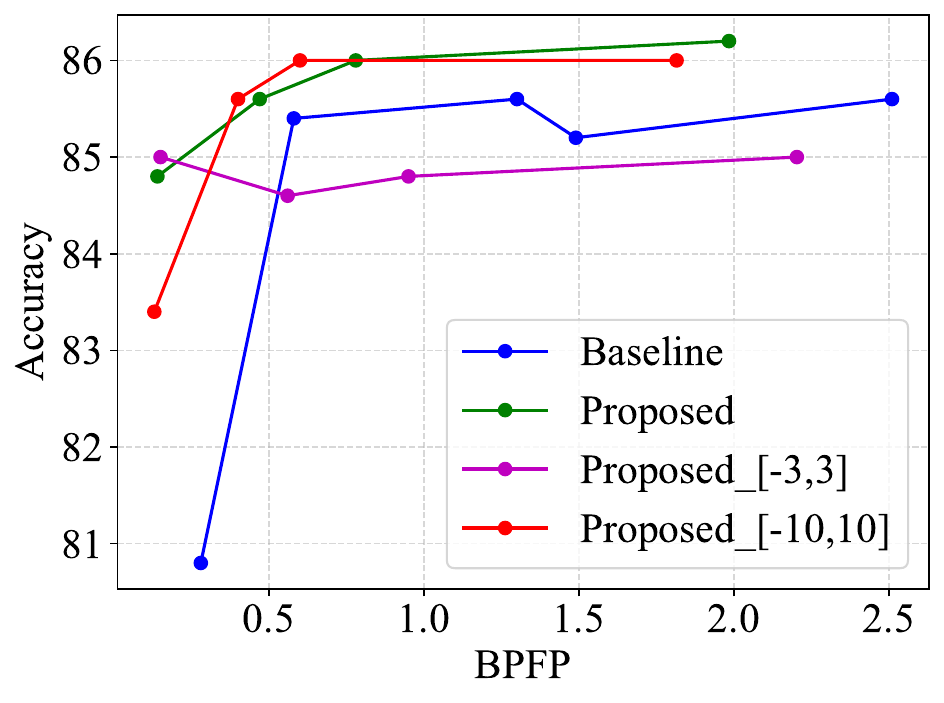}
        \centerline{(a)}\medskip
    \end{minipage}
    \begin{minipage}[b]{0.35\linewidth}
        \centering
        \includegraphics[width=\linewidth]{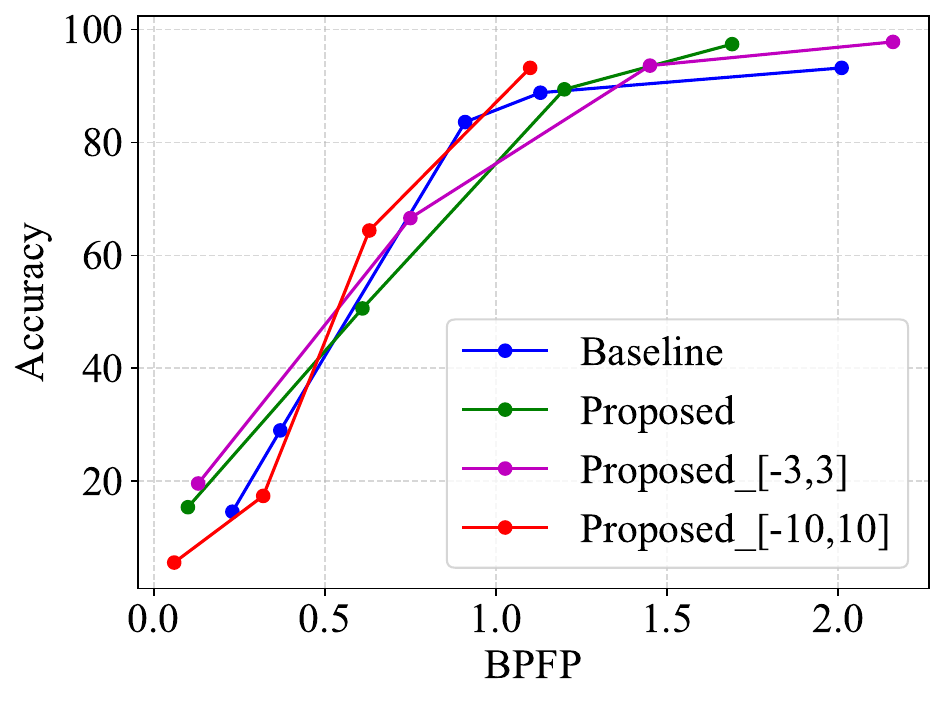}
        \centerline{(b)}\medskip
    \end{minipage}
    \caption{Rate-accuracy performance comparison between the baseline and the proposed methods. (a) ResNet features. (b) DINOv2 features.}
    \label{fig_rd_comparison}
\end{figure*}
\begin{figure}[htb]
    \begin{minipage}[b]{0.49\linewidth}
        \centering
        \includegraphics[width=\linewidth]{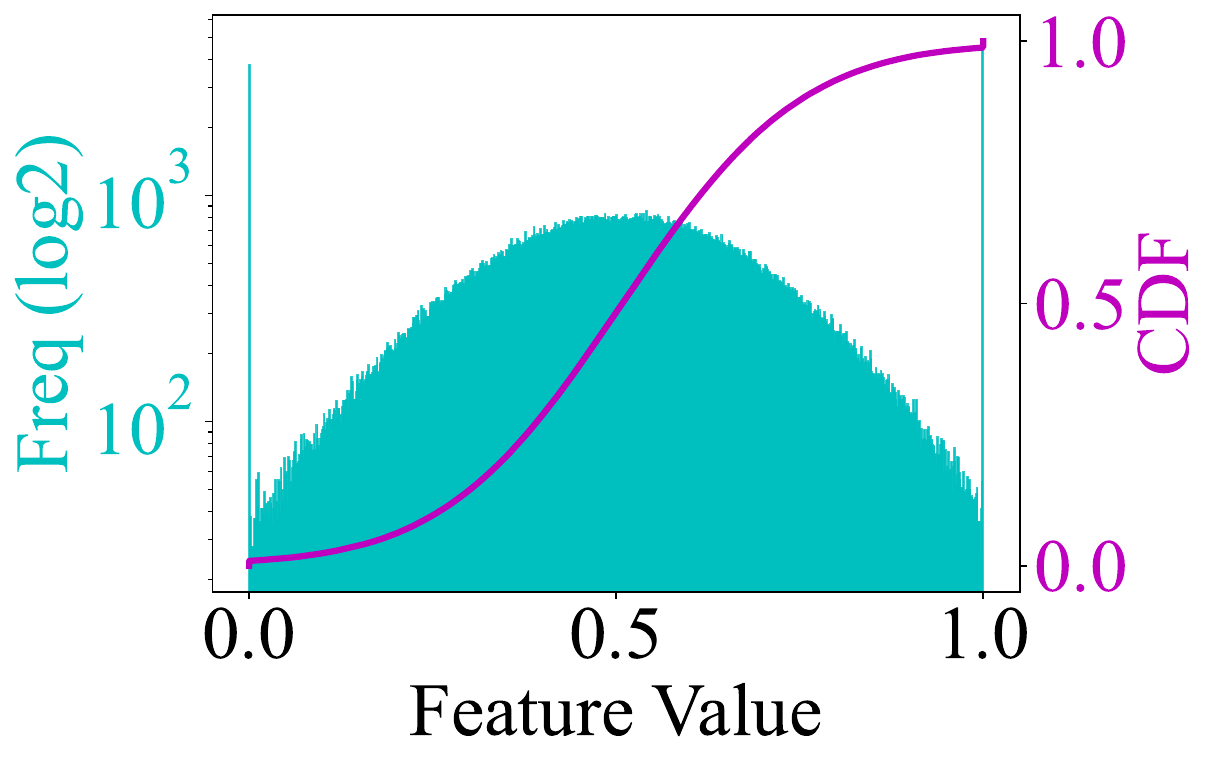}
        \centerline{(a)}\medskip
    \end{minipage}
    \begin{minipage}[b]{0.49\linewidth}
        \centering
        \includegraphics[width=\linewidth]{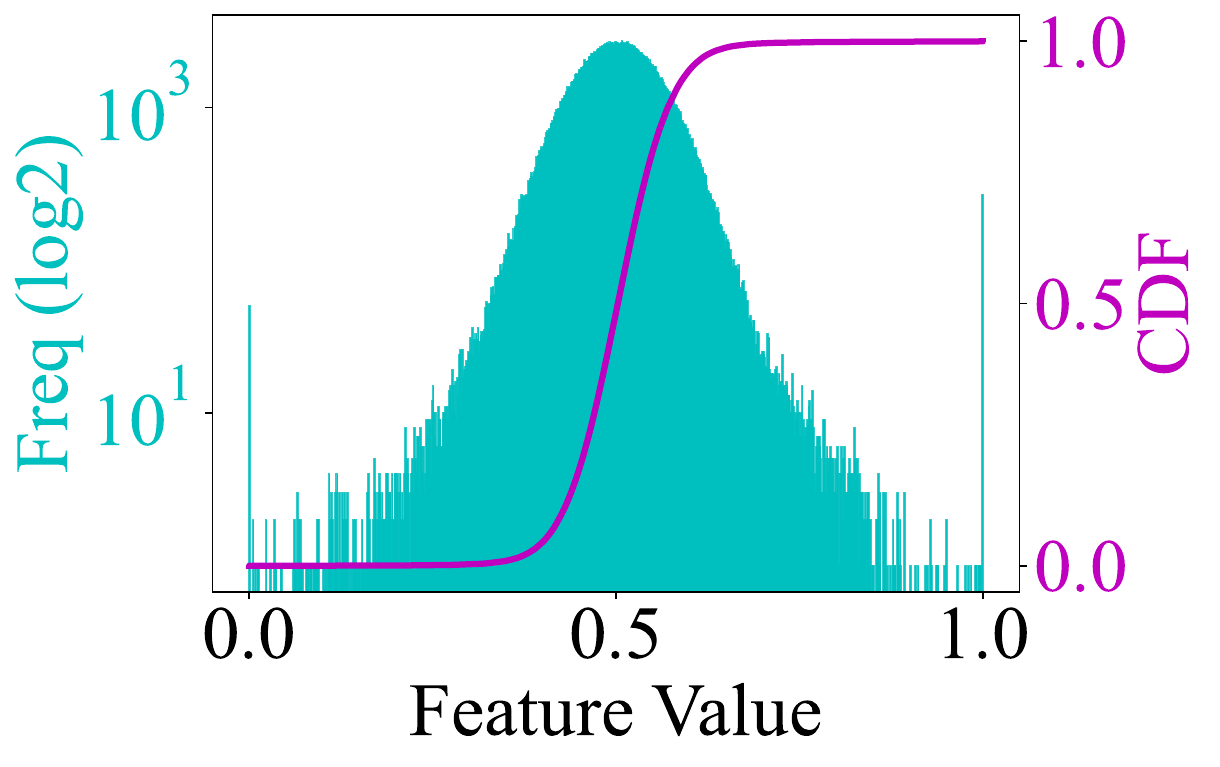}
        \centerline{(b)}\medskip
    \end{minipage}
    \caption{Visualization of standard normalized DINOv2 features truncated to [-3,3] (a) and [-10,10] (b).}
    \label{fig_ablation_value}
\end{figure}
\begin{figure}[htb]
    \begin{minipage}[b]{0.49\linewidth}
        \centering
        \includegraphics[width=\linewidth]{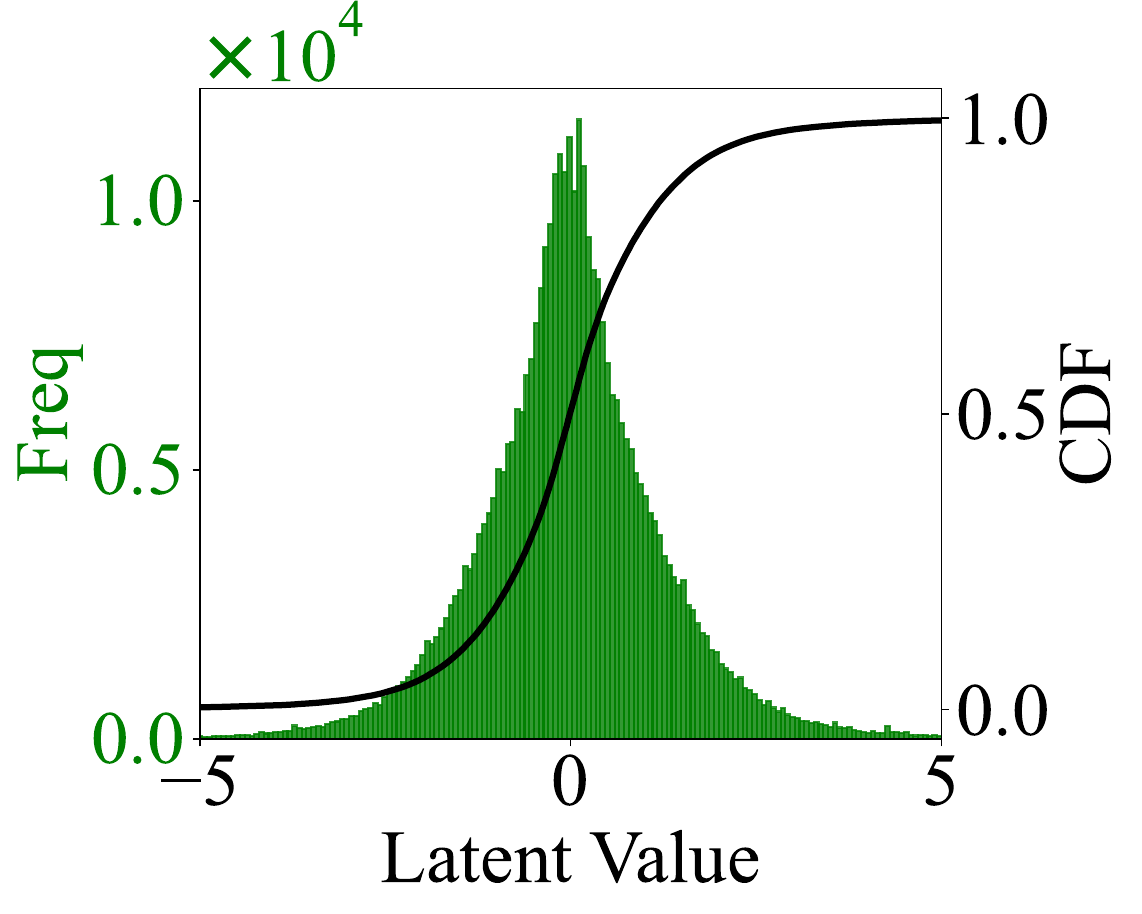}
        \centerline{(a)}\medskip
    \end{minipage}
    \begin{minipage}[b]{0.49\linewidth}
        \centering
        \includegraphics[width=\linewidth]{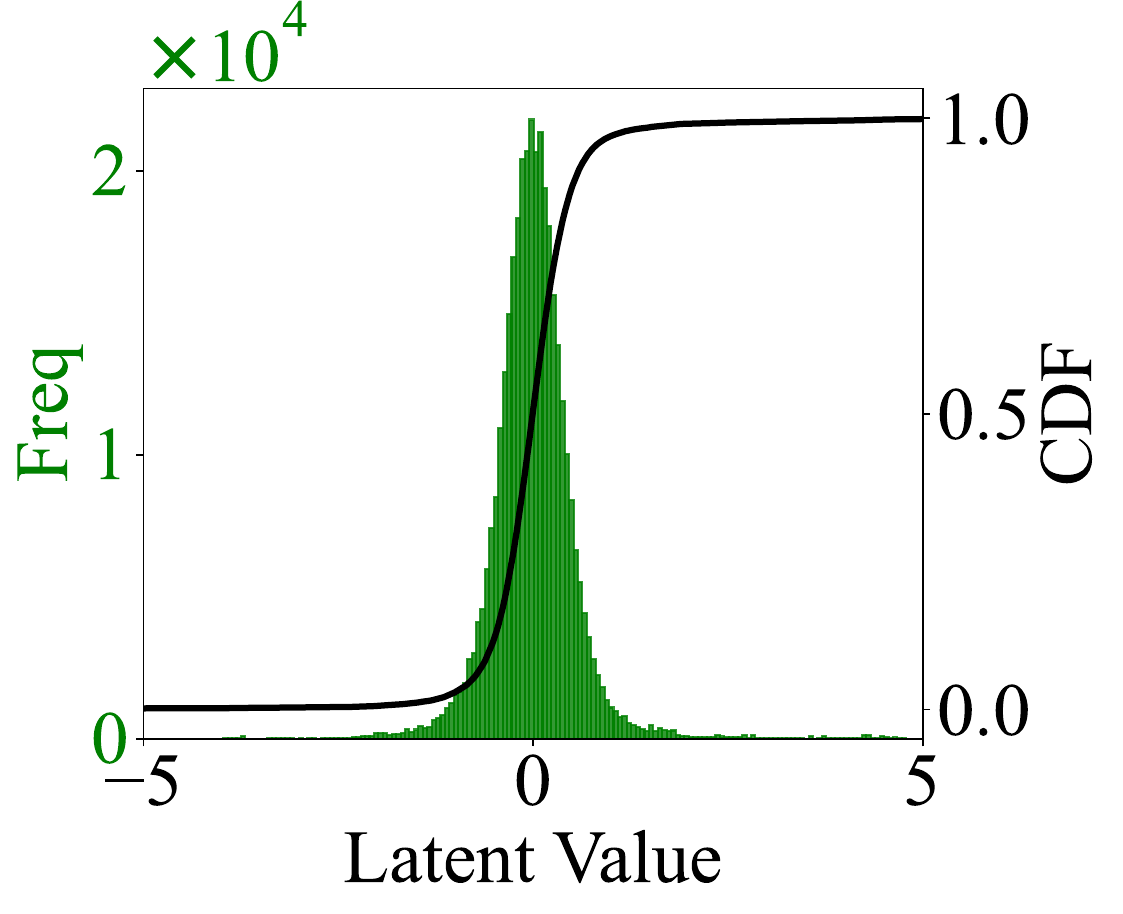}
        \centerline{(b)}\medskip
    \end{minipage}
    \caption{Visualization of latent representations obtained from DINOv2 features truncated to [-3,3] (a) and [-10,10] (b).}
    \label{fig_ablation_latent}
\end{figure}

\section{Experiments}
\label{sec_experiments}
\subsection{Training Details}
We construct our training set by randomly selecting 10,000 images from ImageNet. These images are processed through both ResNet50 and DINOv2 to generate the original features, which are then preprocessed using our proposed truncation and normalization methods.

To achieve different rate-distortion trade-offs, we set lambda values to 0.001, 0.003, 0.005, and 0.01. The corresponding training data sampling ratios are set to 1:1, 1:2, 1:3, and 1:5, respectively. During training, we use 100 DINOv2 features and 100 ResNet features as the validation dataset. 
In our training, we set the initial learning rate to 0.0001 for all bitrates. The learning rate is scheduled by the ReduceLROnPlateau algorithm. We train the model until the learning rate reaches 1e-8. The model with the lowest validation loss is selected to perform feature coding.

\subsection{Evaluation Setup}
Existing feature coding methods for CNN features are incompatible with 2D DINOv2 features as they require 3D inputs. We therefore compare our method against the large model feature coding approach proposed in \cite{gao2024feature}.

We follow the same test conditions and take the classification dataset proposed in \cite{gao2024feature} as our test dataset. The classification accuracy for the original ResNet features and DINOv2 features are 86.4\% and 100\%, respectively. 

\subsection{Rate-Accuracy Performance Analysis}
Figure \ref{fig_rd_comparison} presents the rate-accuracy performance comparison between our proposed method and the baseline. 
For ResNet features, both methods employ shifted normalization to ensure fair comparison. Our method demonstrates consistent accuracy improvements across all bitrates, particularly notable at low bitrates where the baseline exhibits significant accuracy degradation due to its inability to effectively learn the ResNet feature distribution. In addition, the proposed method converges earlier and shows higher stability at high bitrates.

For DINOv2 features, the proposed method achieves higher accuracy at both low and high bitrates, though it exhibits marginally lower performance than the baseline at medium bitrates. Given that both methods utilize identical preprocessing steps and Hyperprior architectures, we attribute our method's overall superior performance to optimized training strategies, particularly in terms of training data sampling and learning rate-based training termination criteria.
\subsection{Ablation on Feature Value Alignment}
In this subsection, we take DINOv2 features as an example to investigate the influence of feature value alignment on the rate-accuracy performance.

We examine three distinct truncation ranges: our proposed method, along with two additional ranges [-3,3] and [-10,10] for comparative analysis. As illustrated in Fig. \ref{fig_ablation_value}, narrower truncation ranges yield flatter feature distributions, while wider ranges produce more concentrated distributions.

The trained codecs' behavior is fundamentally shaped by the training data distribution. Altering the truncation range significantly affects the resulting latent value distributions, as shown in Fig. \ref{fig_ablation_latent}. Wider truncation ranges lead to latent values that cluster more densely around zero, which simplifies probability estimation during entropy coding and consequently reduces bitrates. This phenomenon is consistently observed for both DINOv2 and ResNet features in Fig. \ref{fig_rd_comparison}.

Truncation range selection also impacts classification accuracy. While larger ranges generally enable better rate-accuracy tradeoffs at moderate-to-high bitrates, their effectiveness diminishes at low bitrates. DINOv2 features demonstrate greater robustness to truncation range variations compared to ResNet features, owing to their inherently more concentrated distributions. 
An overly constrained truncation range leads to significant semantic loss in the original features, resulting in substantially degraded accuracy. This effect is particularly evident in ResNet features, where even with abundant bitrate allocation, the [-3,3] truncation range fails to achieve satisfactory accuracy levels. 

These findings demonstrate that careful optimization of the truncation range offers an effective mechanism for balancing rate-accuracy performance in feature coding. 
The demonstrated relationship between truncation range and coding efficiency provides valuable insights that could guide future research in this field.

\section{Conclusion}
In this paper, we presented a cross-architecture universal feature coding method capable of compressing features from both CNNs and Transformers. We proposed a two-step alignment strategy to unify heterogeneous feature representations. 
Experimental results demonstrate that our method achieves superior rate-accuracy performance compared to the existing architecture-specific baseline. 
This study highlights the feasibility and advantages of universal feature coding in heterogeneous model environments. 
Future work will explore extending the proposed method to other modalities and tasks such as segmentation.


\bibliographystyle{IEEEbib}
\bibliography{strings,refs}

\end{document}